\documentclass{article}

    \usepackage[preprint]{neurips_2025}

\usepackage[utf8]{inputenc} %
\usepackage[T1]{fontenc}    %
\usepackage{hyperref}       %
\usepackage[capitalize,noabbrev,nameinlink]{cleveref} %
\usepackage{url}            %
\usepackage{booktabs}       %
\usepackage{amsfonts}       %
\usepackage{nicefrac}       %
\usepackage{microtype}      %
\usepackage[table]{xcolor}         %
\usepackage{soul}
\usepackage{enumitem}
\usepackage{tabularx}  %

\usepackage{todonotes}
\usepackage[normalem]{ulem}

\makeatletter
\newcommand*\iftodonotes{\if@todonotes@disabled\expandafter\@secondoftwo\else\expandafter\@firstoftwo\fi}  %
\makeatother

\newcommand{\authorspace}{\hspace{2em}}
\newcommand{\up}[1]{\textnormal{\textsuperscript{#1}}}

\title{Build the web for agents, not agents for the web}

\author{%
  Xing Han Lù \authorspace
  Gaurav Kamath \authorspace
  Marius Mosbach\up{$\dagger$} \authorspace
  Siva Reddy\up{$\dagger$}\\
  \\
  McGill University\\
  Mila -- Quebec AI Institute\\
  \up{$\dagger$}Equal Advising~~
}

\begin{document}

\maketitle

\begin{abstract}

Recent advancements in Large Language Models (LLMs) and multimodal counterparts have spurred significant interest in developing web agents\textemdash AI systems capable of autonomously navigating and completing tasks within web environments. 
While holding tremendous promise for automating complex web interactions, current approaches face substantial challenges due to the fundamental mismatch between human-designed interfaces and LLM capabilities. 
Current methods struggle with the inherent complexity of web inputs, whether processing massive DOM trees, relying on screenshots augmented with additional information, or bypassing the user interface entirely through API interactions. 
This position paper advocates for a paradigm shift in web agent research: rather than forcing web agents to adapt to interfaces designed for humans, we should develop a new interaction paradigm specifically optimized for agentic capabilities.
To this end, we introduce the concept of an \textit{Agentic Web Interface (AWI)}, an interface specifically designed for agents to navigate a website. We establish six guiding principles for AWI design, emphasizing safety, efficiency, and standardization, to account for the interests of all primary stakeholders.
This reframing aims to overcome fundamental limitations of existing interfaces, paving the way for more efficient, reliable, and transparent web agent design, which will be a collaborative effort involving the broader ML community.

\end{abstract}

\section{Introduction}
\label{sec:introduction}

Rapid advancements in the capabilities of large language models (LLMs), including multimodal models, have resulted in an increased interest in designing \textit{digital agents}: artificial intelligence systems that, unlike conventional LLMs, can not only respond to user input, but also execute a range of actions directly, from booking flight tickets to sending emails. 
For example, if a user requests to book a flight from Montreal to Boston, an LLM may suggest options and provide approximate durations, whereas a digital agent can actually complete the booking on the user's behalf, and have the booking details sent to the user's inbox.
To this end, recent works have focused on the problem of designing agents for different type of interfaces, including web APIs~\citep{schick2023toolformerlanguagemodelsteach,qin2023tool}, mobile UIs~\citep{rawles2023androidinthewild,rawles2024androidworld}, desktop UIs~\citep{xie2024osworld,qin2025ui}, and web browsers~\citep{miniwob++,nakano2022webgptbrowserassistedquestionansweringhuman,gur2023real,he2024webvoyager}. 
The focus of this position paper is on one specific type of digital agents that specializes on navigating websites through mediums like the web browser; we refer to them to as \textit{web agents}.

Acquiring this autonomy unlocks crucial capabilities in web agents that basic chat assistants lack, enabling users to complete web-related tasks that would otherwise not be possible with standard LLM assistants.
One clear advantage is their ability to execute user-defined tasks independently, whether these involve making online purchases or sending messages.
On top of these novel capabilities, web agents also offer more response-adaptability to user inputs than standard LLM-based assistants, due to their ability to interact with online environments.
For example, web agents can address user queries by directly accessing and interacting with websites, making them more suitable for situations where some desired knowledge is not present in text format, is frequently updated, or requires explicit online interaction (such as asking online vendors for price estimates).

Current approaches to building web agents range from zero-shot prompting of LLMs through visual grounding \citep{zheng2024gpt} to finetuning-based approaches that leverage planning \citep{gur2023real,sodhi2024step}, multimodality \citep{shaw2023pixels,furuta2023multimodal}, and reinforcement learning \citep{miniwob++,qi2025webrltrainingllmweb}.
Across all approaches, however, browser states are presented to the agent (i) as screenshots of the browser; (ii) via the underlying web page, as a tree structure refined from the HTML elements present on the page; or (iii) through a combination of the two.
These means of presenting information to the web agent, however, face fundamental drawbacks.
Screenshots do not directly provide agents with all comprehensive webpage information that may be required (for instance, elements in a collapsed dropdown menu will not be visible), while providing web agents full DOM trees is often inefficient and computationally expensive, as these can include details that are irrelevant to the user, and used only to render the webpage.
Similarly, web APIs\textemdash which may not face these representational issues\textemdash encounter limitations in other areas, due to their developer-oriented design.  
Additionally, in both cases, safety concerns arise, given that web agents may have access to sensitive personal information required to execute actions (e.g., payment and login information).

While the approaches discussed above are valid approaches for building web agents, their limitations point to a deeper way in which these approaches are ultimately misguided: they involve trying to build web agents that interact with the web the same way that humans do.
More specifically, the tools that agents are trained to interact with\textemdash e.g., the web browser, or web APIs\textemdash are designed for human users and developers, and not for agentic AI systems.
In this position paper, we argue that agents should \textit{not} be forced to interact with the web in the same way as human users.
\textbf{Beyond designing web agents for human web interfaces, we argue that the research community should be deeply involved in the design of web interfaces for web agents.}

To this end, we present the concept of the \textbf{\textit{Agentic Web Interface}} (\textbf{AWI}): a new type of interface specifically designed to be used by web agents. This mirrors how user interfaces are designed to be used by humans, who do not have the same requirements as web agents. Given the nature of this work, we explicitly do \textit{not} provide a prototype or specific implementation details, as we believe designing AWIs will be a collective and iterative effort involving a broad range of stakeholders.

We structure this paper as follows: in \Cref{sec:background}, we describe current approaches to designing web agents and highlight limitations and safety concerns.
We then introduce AWIs in \Cref{sec:position}, discuss how they can be designed to solve the current issues faced by web agents, and offer guiding principles that should inform their design.
Next, in \Cref{sec:impact_role_of_ML}, we argue that this effort is relevant to and requires contributions from the broader ML community. Finally, we discuss how AWIs differ from the model context protocol (MCP) in \Cref{sec:another_protocol} and summarize our position in \Cref{sec:conclusion}.

\section{How are web agents currently designed?}
\label{sec:background}

Web agents are designed to perform goal-directed tasks through programmatic interaction with web interfaces. To enable this, recent implementations utilize LLMs as a core component of these agents \citep{gur2023real,zheng2024gpt,drouin2024workarena,boisvert2025workarenacompositionalplanningreasoningbased}; however, they are only a part of a bigger system, which interacts with websites through a browser.
Following \citet{zhou_webarena_2024}, we define web agents within a sequential decision-making framework where, given a task described as a natural language intent $i$, e.g., \textit{find all white shoes in size 10 and add them to my wishlist}, the agent issues an action $a_t \in \mathcal{A}$ at time step $t$ according to a policy $\pi(a_t | i, o_t, a_{1}, o_{1})$. 
This policy conditions on the intent $i$, the current observation $o_t \in \mathcal{O}$, the action history $a_{1}$, and the observation history $o_{1}$. 
The execution of action $a_t$ in the environment results in a state transition to $s_{t+1} \in \mathcal{S}$ with a corresponding observation $o_{t+1} \in \mathcal{O}$.
The action space $\mathcal{A}$ of web agents includes operations analogous to human web interactions: element selection (clicking), text input, URL navigation, and browser tab management. 
These actions are executed via programmatic automation frameworks such as Playwright\footnote{\url{https://playwright.dev}}, which return updated browser states after each agent intervention. 
The effectiveness of task completion can then be evaluated through a reward function $r(a_{1}, s_{1})$ that assesses whether the final state satisfies the intent criteria. The reward function be an automatic evaluator that programmatically determine success of a final state based on a human-defined rules \citep{zhou_webarena_2024,koh_visualwebarena_2024,drouin2024workarena}, or a specialized reward model that can determine the success of a trajectory without having access to the underlying environment or rules \citep{murty2024bagelbootstrappingagentsguiding,l2025agentrewardbench,xue2025illusionprogressassessingcurrent}. 

In this section, we provide an overview of web agents by examining two approaches: agents that solely rely on web browser UIs,  (\S\ref{sec:background_browser_based_agents}), and agents that additionally incorporate web APIs (\S\ref{sec:background_tool_use}). 
We briefly cover each category and highlight the limitations they each face as a result of having to interact with interfaces designed for humans.

\subsection{Browser-based agents}
\label{sec:background_browser_based_agents}
Web agents that rely on the browser's UI may leverage visual representations from the screen \citep{miniwob,shaw2023pixels}, the Document Object Model (DOM) tree \citep{deng2023mind2web}, the accessibility tree \citep{zhou_webarena_2024}, or a combination of them \citep[e.g.,][]{furuta2023multimodal, lu2024weblinx,boisvert2025workarenacompositionalplanningreasoningbased,gou2025navigatingdigitalworldhumans}.
In addition to the raw visual and textual inputs, additional annotations like bounding boxes are sometimes added onto the screenshots \citep[e.g.,][]{yang2023set,koh_visualwebarena_2024,dechezelles2025browsergymecosystemwebagent}.  
This multimodal representation allows a holistic representation of the webpage, with the DOM showing the structure of the webpage and screenshots showing images that are not displayed inside the DOM. This paradigm enables the development of agents capable of predicting and executing browser actions, such as element selection, text input, URL navigation, and tab management operations.

\paragraph{Representational limitations}
Browser-based web agents face limitations imposed by both DOM-based and screenshot-based representations.
Screenshots provide token-efficient representations of browser state, but lack comprehensive DOM information that may be visually occluded \textemdash i.e., not rendered or hidden inside a dropdown.
On the other hand, DOM-based representations, despite usually containing the most decision-relevant information, are extremely inefficient due to excessive structural tokens and supplementary attributes (such as server-side identifiers).
This inefficiency in turn leads to higher computational and operational costs when using LLMs; for instance, with DOM trees potentially exceeding 1M tokens \citep{dechezelles2025browsergymecosystemwebagent}, deploying a GPT-4.1-based agent for a single 20-step task could cost roughly \$40\footnote{This estimate assumes \$2 per million input tokens, and may increase for reasoning models.}.
While previous research has attempted various mitigation strategies \citep[e.g.,][]{deng2023mind2web, tiwary2024contextactionanalysisimpact, lu2024weblinx}, these solutions typically do not generalize well across websites and novel task scenarios. 

\paragraph{Resource challenges and defensive designs}
Beyond representational limitations, the proliferation of browser-based agents introduces additional resource challenges.
Serving web pages can often be resource-intensive, and repeated rendering by browser automation tools (e.g., for web crawling) can lead to a strain on web infrastructure, leading to performance degradation for end users.
As web agents begin to constitute a larger and larger proportion of web traffic, we can expect this resource consumption issue to increase in severity.
In response to such challenges, website operators implement defense mechanisms like CAPTCHA; but as agent capabilities have become increasingly sophisticated, CAPTCHA systems have similarly become increasingly complex, creating accessibility barriers for legitimate human users. Although exclusion protocols like \textit{robots.txt} could be enforced by regulatory bodies for major providers, they will penalize agents used as an assistive technology indistinguishably from agents built for web crawling.
More generally, challenges arise in distinguishing between malicious automation and beneficial human-in-the-loop applications, leading to difficulties in web resource management.

\paragraph{Safety and privacy concerns}
Since they are integrated within the browser, browser-based web agents introduce major safety and privacy concerns. 
If the web agent has access to the user's personal accounts, as well as other sensitive information stored in the browser\textemdash such as passwords and credit card information\textemdash agents that lack adequate safeguards may cause severe harm to users through their actions.
A web agent may, for example, use the user's personal information to send harmful messages or make unauthorized online purchases \citep{tur2025safearenaevaluatingsafetyautonomous,boisvert2025doomarenaframeworktestingai}.

\subsection{API-based hybrid agents}
\label{sec:background_tool_use}
Agents that use tools like web APIs offer high potential benefits to users, particularly when considering tools like Deep Research \citep{deep_research_openai}, that can themselves inform agent behavior.
\citet{song2025browsingapibasedwebagents} defines a \textit{hybrid agent} that extends the browser-based agent by giving it access to the underlying web API of the websites. 
Their agent alternates between API calls and browser interactions to complete a web navigation task from the WebArena benchmark \citep{zhou_webarena_2024}. 
While the incorporation of web APIs by hybrid agents opens new possibilities, it also raises questions about fundamental limitations of web APIs in agentic contexts. 
Most crucially, it begs the question: can web agents solely rely on the internal web API offered by websites?
We argue that besides the previously discussed limitations that solely browser-based agents face, this approach will face limitations and safety concerns specific to web APIs used to power websites.

\paragraph{Limitations of web APIs} 
API-based agents are limited by the range of actions offered by the web APIs, which are far narrower than the range of actions offered by webpage UIs.
Similarly, web APIs are typically not designed to directly manipulate stateful objects, such as sorting a list of products on a online shopping website. This limits the potential actions that an API-based agent could utilize. One could add state-centric actions, but such an endeavour would require substantial refactoring effort from the developers, which would solely benefit API-calling agents.
Additionally, web APIs tend to be heavily controlled by the developers, and frequent usage of the APIs in a short period can lead to request denials.
These factors hold back the effectiveness of incorporating web APIs into web agents, as they have to fall back to using the UI when an action cannot be completed using the web API, which face the limitations discussed in \Cref{sec:background_browser_based_agents}.

\paragraph{Safety concerns}
Internal web APIs are designed to strictly communicate with browsers, as opposed to external APIs that are designed as a service. If used directly, internal APIs may pose security risks, as agents may lack essential safeguards to avoid unintended side effects \citep{levy2024stwebagentbenchbenchmarkevaluatingsafety,boisvert2025doomarenaframeworktestingai} and are prone to follow malicious and harmful instructions \citep{andriushchenko2025agentharmbenchmarkmeasuringharmfulness,tur2025safearenaevaluatingsafetyautonomous}. More concretely, guardrails like password prompts and two-factor authentication can be bypassed if the agent directly communicates with the web API through an elevated API key. This could lead to high API usage if uncontrolled, which could results in unexpectedly high expenses.

\section{Rethinking how agents interact with the web}
\label{sec:position}

As we have argued, the two primary entry points for agents to interact with the web\textemdash UIs and tools (e.g., web APIs)\textemdash are not optimized for use by agents, causing downstream issues with agent deployment.
In order to address these limitations, we argue that the agent research community should rethink how agents interact with the web. 
To this end, we believe that ML researchers should work together to design unified interfaces for web agents to optimally access web content.
We refer to these as \textit{Agentic Web Interfaces}, and believe they are essential for web agent performance.

Whereas APIs are targeted at developers and browser UIs are designed for users, AWIs should be built for agents to assist  users in a interactive setting or to complete user-defined tasks autonomously.
In this section, we examine how AWIs approach the limitations faced by existing interfaces (\S\ref{sec:awis_addresses_issues}), offer guiding principles for designing AWIs (\S\ref{sec:guiding_principles}), and then conclude by presenting more concrete suggestions for AWI design (\S\ref{sec:concrete_suggestions_for_awis}).

\subsection{Addressing issues faced by agents through AWIs}
\label{sec:awis_addresses_issues}

Existing work has primarily focused on designing agents around existing web interfaces like APIs and browser UIs, which are faced with the limitations discussed in \Cref{sec:background}. 
Such issues are unlikely to be directly solved by API and UI developers, as their primary focus is to build interfaces for humans. 
At the same time, designing workarounds for existing interfaces would require a substantial and decentralized effort from individual groups of researchers, which may result in mutually exclusive solutions that may eventually be superseded after major updates to an API or UI. 
For instance, a popular website may see its client-side source code completely rewritten from an older JavaScript framework to a more recent one, which would completely change the observations received by the agent, while keeping the same UI. 
Similarly, a website may go through a complete visual redesign, but the underlying architecture may remain the same.
We therefore believe that the long-term solution is to design AWIs: web interfaces that are specifically designed for use by agents.

Developing AWIs will directly fix many of the issues highlighted above that agents face.
While the limited range of actions offered by most current web APIs means that tool use agents face constraints in how they can interact with online content, AWIs would allow for action spaces to be tailored to desired agent use.
For example, an AWI may deliberately allow a web agent to execute more actions than the web API allows for, but fewer actions than a human user is allowed to execute.
Similarly, AWIs would address the representational limitations and resource challenges faced by browser-based web agents.
Designing dedicated web interfaces for agents would involve presenting them with tailored browser state representations that neither contain excessive, superfluous details (like the raw website DOM tree), nor lack comprehensive DOM information (like screenshots).
This higher efficiency would significantly reduce the computational cost of web agents.
On the other hand, dedicated AWIs would address concerns of web resource management, as they could involve agentic task queues (see \Cref{sec:concrete_suggestions_for_awis}) that handle the traffic of agents while consistently ensuring accessibility and adequate resources for human users. 
This effectively provides an incentive to website owners to implement AWIs, as they would avoid development effort towards managing resources due to higher traffic from agents. 
Additionally, the implementation of AWIs can be potentially abstracted through the use of coding agents \citep{yang2024swe}, which could reduce the barrier for adopting AWIs.
Finally, AWIs would address many of the safety concerns entailed by web agents.
By controlling the information that agents have access to, and carefully limiting its range of executable actions, AWIs enable vital safeguards for web agent behavior.

\subsection{Guiding principles for designing AWIs}
\label{sec:guiding_principles}

We believe the design of AWIs should be iterative and involve a wide range of stakeholders\textemdash from ML researchers to website developers and end users\textemdash so that AWIs enable agents to navigate the web in ways that align with the interests of all stakeholders. 
Consequently, we do not present any single precise blueprint for designing AWIs, as optimal designs will likely require experimenting with several alternatives to identify what works best for a broad range of websites and users.
Instead, we provide a list of guiding principles that we feel are crucial to design good AWIs:

\begin{enumerate}[leftmargin=3em]
    \item \textbf{Standardized}: AWIs should follow standards that clearly define the structure of the interface and the action space supported. They would need to be designed by a working group of experts across ML disciplines, who ensure that they are compatible with various agent designs. 
    \item \textbf{Human-centric}: AWIs should be designed to be used by agents that benefit human users. They should always preserve human agency, safety and privacy. For example, a human should be able to pause an ongoing trajectory and steer the agent by requesting changes to the task, which should work seamlessly with any AWI.
    \item \textbf{Safe}: AWIs should be designed to be safely used by agents, and should have defense mechanisms against malicious agents, malicious content, and catastrophic failures during trajectories; this can be accomplished through proper access controls, guardrails and privacy-preserving methods, which will ensure the safety of the users and websites.
    \item \textbf{Optimal representations}: AWIs should produce efficient representations when provided as observations to web agents. They should contain enough information for the agent to optimally solve the tasks and should not include information that the agent will not need. Moreover, agents should be able to specify exactly the information type (e.g., images) and parameters (e.g., resolution and compression level).
    \item \textbf{Efficient to host}: AWIs should avoid increasing the total computational load of a website. Since agents may be deployed in autonomous settings, or may attempt to navigate multiple websites at the same time, the total traffic will increase as web agents become more widespread. Thus, the efficiency will be crucial to ensure the scalability of web trajectories inside real web environments without increasing hosting costs for website owners.
    \item \textbf{Developer-friendly}: Whereas researchers may be the ones designing AWIs, the developers of the websites will ultimately implement, deploy and maintain them. AWIs will need to work seamlessly with the architecture and infrastructure of the website, and ensure that it does not disrupt the hosting and reliability of the web services.
\end{enumerate}

\subsection{Concrete suggestions for AWIs}
\label{sec:concrete_suggestions_for_awis}
Given the guiding principles mentioned above, we provide several suggestions that could be incorporated into future AWI designs. 
Our recommendations may be not be universally relevant to all use-cases, but we believe they will be a helpful initial step towards more detailed blueprints.

\paragraph{Unified higher-level actions}
In addition to a more efficient representation of the trajectory state, the interface can further augment the action space by abstracting primitive elements, allowing for unified high-level actions. In existing action spaces for web tasks, such as the one defined by BrowserGym \citep{dechezelles2025browsergymecosystemwebagent}, we are already able to find actions like \texttt{goto}, which combines the actions of selecting the address bar, inserting an URL address, and pressing \textit{Enter} to load the webpage. We can view them as high-level functions, which can take higher-level inputs and compose primitive functions (click, type, etc.) that can induce a desired outcome like opening a webpage by its URL address. 
By unifying the action space within the specifications of the agentic web interface, the same higher-level actions can be used across different websites with the same outcome, without having to worry about the internal implementation of the website. 
\paragraph{Compatibility with user interfaces}
To enable the use of web agents on UIs for humans, developers can decide to design AWIs to be compatible with traditional browsers. To accomplish this, actions on AWIs could be executable on the UIs, and the state of AWIs could match updates to the state of the UI. To this end, AWI developers can build a bidirectional translation tool that allows converting actions between AWIs and UIs, which would allow the UI to match the state of the AWI; this could be accomplished through tools like Playwright or Selenium\footnote{\url{https://selenium.dev}}. The resulting tool can be shared among the community, allowing faster developments of UI-compatible AWIs.

\paragraph{Access control for agents}

Given agents would navigate the website through an AWI, we would need to define \textit{Access Control Lists} \citep{rfc4949} specifically for agents. The agent should have restricted access to sensitive user information, such as their credit card information; to ensure such information can be used, they can use privacy-preserving password managers. Moreover, the agent could require the explicit confirmation of the user before performing potentially destructive actions, such as deleting an account. 
Changing the access of the agent would enable greater level of security, which is an area of research previously explored \citep{levy2024stwebagentbenchbenchmarkevaluatingsafety}, and would allow the users to set their preferred permission levels for web agents that use their accounts.
To differentiate the user from the agent, the website could request passkeys or biometric authentication, avoiding CAPTCHAs.

\paragraph{Progressive information transfer}

Many web tasks require a high amount of bandwidth and can be compute-intensive.
For instance, a gallery of high-resolution images could take over 100MB to be sent to the user, and would require expensive animations to display. However, the agent would not need the high-resolution images nor the animations to solve a task.
Instead, AWIs can constrain the information to be progressively sent to the agent in more efficient ways.
In our example, the AWI could send an initial portion of the images resized to an optimal size or as embeddings. This would save a substantial amount of bandwidth, thus leading to reduced costs and latency.

\paragraph{Agentic task queues}

Designing task queues specifically for AWIs would allow the website developer to set a maximum number of concurrent agents to access and navigate a website, with subsequent agents having to wait until a slot frees up before starting their navigation. The queue could distribute the access across the day to avoid usage spikes on certain times of the day, enabling a greater number of agents to be included in a queue for a website. As a result, human users can use the website as they usually would, without worrying about latency caused by usage spikes.

\section{Why must the broader ML community be involved?}
\label{sec:impact_role_of_ML}

Our position results not merely from an engineering convenience, but instead from a strong belief that a fundamental reorientation of web agent research is needed.
By shifting focus from navigating interfaces built for humans to building agent-specific communication standards, we can accelerate progress toward genuinely intelligent web agents while also establishing more rigorous foundations for measuring and advancing agent capabilities.
As a result, we believe that implementing AWIs is of great relevance not only to those working on downstream applications of web agents, but rather the ML community at large.
Below, we highlight the relevance of agentic web interfaces to various ML communities, in order to underscore both the importance of AWIs for their research, and the need for their involvement in designing AWIs.
We show how each principle and suggestion corresponds to one or more disciplines in \Cref{tab:main_table}.

\paragraph{Human-centric AI (HCAI)}
An AWI will enable further opportunity to design personalized agents\textemdash agents that are specifically tailored to a user's individual preferences \citep[e.g.,][]{cai2025large}.
Additionally, designing AWIs to allow both UI-based and chat-based web navigation will help improve the experience in scenarios where the user has access and knows how to use a UI, as well as a scenario where the user wishes to avoid using the UI, instead deferring the task to the web agent. 
Rather than building web agents that replace users, HCAI research would instead help guide the design of interoperability between AWI and the traditional UI in a way that augments users, who may expect different levels of autonomy from agents based on their preferences \citep{Ge_2024}.
Thus, having HCAI researchers partake in the design process would allow AWIs to be implemented in a way that will benefit a broader range of user goals and preferences.

\paragraph{AI Safety}
Recent works on agent safety \citep{levy2024stwebagentbenchbenchmarkevaluatingsafety,tur2025safearenaevaluatingsafetyautonomous,andriushchenko2025agentharmbenchmarkmeasuringharmfulness} highlight the substantial effort that will be required to design agents that are not only capable, but also safe. As the adoption of web agents increases, they will used in progressively more critical and sensitive scenarios.
The AI safety research community will therefore play a crucial role in guiding the design and development of AWIs, enabling safer out-of-the-box web environments for web agents to navigate.
Notably, AI safety researchers can design AWIs to be robust to, prompt-level \citep{liao2025redteamcuarealisticadversarialtesting}, HTML-level \citep{boisvert2025doomarenaframeworktestingai} or vision-level attacks \citep{wu2024adversarial}, and to carefully choose ACLs based on the nature of the task and the sensitivity level of the website.

\paragraph{Natural Language Processing (NLP)}
Several topics within NLP can leverage AWIs to build more capable agents. For example, to achieve an optimal representation, AWIs could use iterative summarization \citep{wan2007towards,yan2011evolutionary,zhang2023summit} to abstract unnecessarily verbose sections of the webpage, while keeping more relevant ones intact. Not only would this help AWIs produce concise representations of a website to enhance the decision making of web agents, they will also provide salient representations that can be stored in a collection, enabling them to be retrieved  in a future task as supporting artifacts. Thus, AWIs could be built with retrieval augmentation \citep{lee2019latentretrievalweaklysupervised, guu2020realmretrievalaugmentedlanguagemodel,lewis2020retrieval} in mind, enabling the use of retrieval methods like dense encoders \citep{karpukhin2020densepassageretrievalopendomain}, late interactions \citep{khattab2020colbertefficienteffectivepassage}, and LLM-based embeddings \citep{behnamghader2024llm2vec}. The retrieval-augmented web agents can leverage the retrieved knowledge or memory segments to complete contextual tasks, which would have otherwise required clarification from the user or resulted in a refusal.

\renewcommand{\arraystretch}{1.65}  %
\begin{table}[t]
\caption{In this table, we enumerate the guiding principles (\S\ref{sec:guiding_principles}) alongside corresponding suggestions (\S\ref{sec:concrete_suggestions_for_awis}) and relevant ML disciplines (\S\ref{sec:impact_role_of_ML}). The guiding principles should inform any concrete implementation steps, and determine which other ML disciplines must be involved.}
\label{tab:main_table}
\rowcolors{2}{white}{black!6}  %

\begin{tabularx}{\textwidth}{>{\hsize=.8\hsize}X | >{\hsize=1.2\hsize}X >{\hsize=1\hsize}X}
\hline
\textbf{Principles (\S\ref{sec:guiding_principles})} & \textbf{Suggestions (\S\ref{sec:concrete_suggestions_for_awis})} & \textbf{ML Disciplines Involved (\S\ref{sec:impact_role_of_ML})} \\
\hline
Standardized & Unified higher-level actions & Generalization, NLP \\
Human-centric & Compatibility with user interfaces & HCAI \\
Safe & Access control for agents & AI Safety \\
Optimally-represented & Progressive information transfer & Generalization, NLP, RL \\
Efficient to host & Progressive information transfer & Planning, RL \\
Developer-friendly & Agentic task queues & \textit{All} \\
\hline
\end{tabularx}
\end{table}

\paragraph{Multimodality}
For researchers working on multimodality, AWIs hold great potential, as they represent an opportunity to boost the multimodal capabilities of agents.
Current screen-based approaches often put the multimodal abilities of agents at a disadvantage, as visual artifacts (such as product images in an online catalogue, or photos in a gallery) tend to be passed to the agent as part of a single snapshot \citep{shaw2023pixels,rawles2024androidworld,zheng2024gpt,dechezelles2025browsergymecosystemwebagent}.
With AWIs, multimodal models could be provided purposefully processed media from a webpage, in formats that benefit their capabilities\textemdash such as resized versions of images that take up too much or too little space in a single screenshot, or truncated versions of long video or audio recordings.
With the help of benchmarks focused on the multimodal capabilities of web agents \citep[e.g.,][]{koh_visualwebarena_2024,jang2024videowebarena,liu2024visualwebbench, liu2024visualagentbench}, research on multimodality can help quantify the effectiveness of different representations of multimodal artifacts for web agents.
In turn, these findings can be incorporated into AWI design, creating a virtuous cycle that enables multimodal research to focus less on adapting models to human interfaces and more on how to represent multimodal inputs.

\paragraph{Reinforcement Learning}
Designing AWIs will have strong consequences for reinforcement learning, as the limitations of current approaches to web agent development do not only extend to the safety and resource allocation concerns discussed already; instead, they cause inefficiencies in the learning process itself.   
Under current approaches, computing a reward signal is unnecessarily complex, requiring elaborate engineering to determine task completion based on UI state \citep{liu2023agentbenchevaluatingllmsagents}.
Instead, a standardized interface design would enable the development of reward functions that can produce consistent success or failure signals.
Similarly problematic is the inconsistent action space that agents must navigate. 
Current approaches require web agents to learn low-level UI actions \citep{CacciaFineTuningWA,patel2024largelanguagemodelsselfimprove,murty2024bagelbootstrappingagentsguiding,murty2025nnetnavunsupervisedlearningbrowser,qin2025ui}, which can be inconsistent across agent frameworks. For instance, BrowserGym's action space \citep{dechezelles2025browsergymecosystemwebagent} defines actions for sending messages to the user and for opening or closing tabs. On the other hand, the concurrent action space proposed by \citep{qin2025ui} does not include any tab-level action and does not allow agents to respond to user, instead defining an action specifically for requesting user intervention.
A unified high-level action standard would supersede action spaces defined by individual research works with a consistent and extensible action space, enabling policy learning to be compatible across different frameworks.

\paragraph{Planning}
AWIs will further enable research on search and exploration -- or, more broadly, planning -- in the context of web agents. 
By strictly interacting with AWIs, agents that leverage planning would be able to create search paths in the sandbox version of a website without overloading it for human users, which has shown promising results in simulated paths \citep{gu2025llmsecretlyworldmodel}. Through the more efficient design and the isolated nature of the interface, we can substantially increase the number of episodes that an agent can complete within the same timeframe, while ensuring that the website remains functional for human users. Consequently, methods like Reflexion \citep{shinn2023reflexionlanguageagentsverbal}, SkillWeaver \citep{Zheng2025SkillWeaverWA} and Tree Search \citep{koh2024treesearchlanguagemodel} can be scaled to achieve better test-time performance, whereas methods like ADaPT \citep{prasad2024adaptasneededdecompositionplanning} can be accelerated to enable real-time usage. Thus, designing AWIs to be efficient for planning-based agents would result in reduced website usage load and bandwidth costs, which could incentivize more website owners to adopt sandboxes for search-based agents.

\paragraph{Generalization}
For researchers focused on task generalization, AWIs provide an opportunity to better study the generalization capabilities of agents.
Many current approaches involve training agents on specific UI interaction patterns \citep{furuta2023multimodal,lai2024autowebglmlargelanguagemodelbased,trabucco2025towards}, which ties agent capabilities to particular interface implementations. This may result in brittle systems that fail when encountering novel UI designs and therefore generalize poorly across websites \citep{Li2024OnTE,lu2024weblinx,pan2024webcanvasbenchmarkingwebagents}.
Training on standardized implementations, however, allows one to decouple task knowledge from interface knowledge, holding the potential for more effective transfer across websites. 
As a result, ML researchers can focus on higher-level generalization problems (such as adapting agents to solve novel tasks) rather than the lower-level problems caused by differences in the DOM trees and UIs between websites.

\section{Is an AWI another communication protocol for LLM agents?}
\label{sec:another_protocol}

Tool use for LLMs is a well researched topic \citep{qin2023tool,schick2023toolformerlanguagemodelsteach}, which paved the way for providers to integrate tool use into popular LLM assistants, such as Search for ChatGPT \citep{chatgpt_search_2024} and Function Calling for Gemini \citep{gemini_function_calling}. 
As adoption grows, a need to standardize interactions between LLM agents (i.e., \textit{hosts}) and tools (i.e., \textit{servers}), led to the introduction of the \textit{Model Context Protocol (MCP)} \citep{mcp_anthropic_2024}. 
The protocol aims to standardize how LLM assistants, such as Claude \citep{anthropic_claude_2023}, communicate with externally-hosted systems like PostgresSQL\footnote{\url{https://www.postgresql.org}} databases, Slack\footnote{\url{https://slack.com}} workspaces, and GitHub\footnote{\url{https://github.com}} accounts. 
The protocol enables LLM assistants to use a \textit{client} to access MCP servers, where they can send a query using the JSON-RPC 2.0 \citep{json_rpc_2} format, and receive a response in the same format. 
By standardizing how information is sent and received, the protocol enables MCP servers to be effortlessly compatible with any MCP-compatible LLM agent, avoiding the need for wrappers around API endpoints provided by web services. 
However, although MCP serves as a major catalyst for developing tool use agents, it fundamentally differs from AWIs, which are \textit{interfaces}\textemdash not a \textit{protocol}\textemdash designed for web agents that navigates stateful representation of websites.
Below, we highlight two key differences between MCP and AWIs, which further motivate the need for the latter.

\paragraph{State tracking}
Since MCP uses JSON-RPC 2.0\textemdash a stateless protocol\textemdash it will not directly support client-side state tracking, thus limiting state-dependent actions. 
For instance, if an agent is tasked to purchase white shoes, it may first request the list of all shoes fitting a certain size, which it can easily receive from an AWI or from a MCP server; this becomes its latest state. 
Given this state, an AWI-based agent can request the list to be sorted by price. 
On the other hand, an MCP client would have to reformulate a query that requests the same list again, this time sorted by price. 
This overhead would lead to substantially higher bandwidth cost (since the entire list needs to be sent again) and slowing down the task completion. 
Unlike MCP servers, AWIs are designed to universally track states, enabling more efficient ways of executing actions that depends on the state of a website.

\paragraph{Standardized interface vs protocol}
Although the communication protocol is standardized, the actual implementation may differ based on the server. 
For instance, an implementation for GitHub\footnote{\url{https://github.com/modelcontextprotocol/servers/tree/2025.5.12/src/github}} may have a function \texttt{get\_file\_contents} that takes as inputs \texttt{owner} and \texttt{repo} and return the content of a file; the same function for GitLab\footnote{\url{https://github.com/modelcontextprotocol/servers/tree/2025.5.12/src/gitlab}} may instead require \texttt{project\_id} as input. 
This difference stems from the flexibility of the protocol, which allows MCP servers to specify their own \textit{methods} and \textit{parameters}. 
On the other hand, AWIs will be standardized across implementations, similar to how a \texttt{FileReader} object in JavaScript will behave the same across different browsers.\\

Overall, Agentic Web Interfaces enable agents to navigate websites, whereas MCP provides a unified approach for LLM assistants to communicate with a wide range of web services. 
That being said, AWIs and MCPs need not be mutually exclusive; AWIs can be designed to communicate with a web service through MCP, whereas an MCP server could access websites through AWIs and a server-side agent, enabling MCP-compatible LLM assistants to navigate websites autonomously. 
Ultimately, we believe that MCP and AWIs will both be crucial for developing more capable tool use and web agents, though, once again, the two are fundamentally different.

\section{Conclusion}
\label{sec:conclusion}

Web agents represent one of the most exciting current areas in AI research, with a high potential for impact on the daily lives of everyday users.
As we have argued, however, research in the field is currently being held back by limitations imposed by interfaces that are built for human users, and not web agents. 
These limitations not only act as a bottleneck for research on web agents, but also obscure their potential capabilities, introduce safety risks, and will lead to major resource challenges.
Consequently, we argue that it is imperative that the broader research community helps design \textit{agentic web interfaces} (AWIs): interfaces designed specifically to be used by web agents.
In this position paper, we highlighted the need for AWIs (\S\ref{sec:background},\S\ref{sec:awis_addresses_issues}), and offered both high-level principles (\S\ref{sec:guiding_principles}) and concrete design suggestions (\S\ref{sec:concrete_suggestions_for_awis}) that we believe should be held in mind when building them.
We underscored the relevance of AWIs to the wider ML community, stressing that designing better interfaces for web agents is not a concern only for web developers and web agent researchers (\S\ref{sec:impact_role_of_ML}), but the research community at large. Finally, we highlight the difference between a protocol like MCP and web interfaces dedicated for agents (\S\ref{sec:another_protocol}).

Given the technical challenges discussed in \Cref{sec:impact_role_of_ML}, the development of web standards requires ML expertise from the outset. 
In our view, the machine learning community should be actively involved in the design of AWIs from their inception\textemdash ensuring that elements to facilitate testing, debugging, and safety are incorporated into the standard. 
Treating these capabilities as first-class considerations, rather than retrofitting them onto systems designed primarily for human interaction, should be one of the main priorities in the design of agentic web interface standards.
Web agents hold the potential for major societal impact; it is important that we design interfaces that reflect this importance.

\section*{Acknowledgement}
\label{sec:acknowledgment}

Xing Han Lù acknowledges the support of the Natural Sciences and Engineering Research Council of Canada (NSERC) [funding reference no. 579403].
Siva Reddy is supported by a Canada CIFAR AI Chair.
We thank Alexandre  Drouin, Alexandre  Lacoste, Christopher Pal, Maxime Gasse, Peter Shaw, Tianbao Xie, Yu Su, and the McGill NLP group members for sharing their feedback and discussing their agreements and disagreements on the position of this paper.

\bibliographystyle{abbrvnat}
\bibliography{references}

\end{document}